\documentclass{article}
\usepackage{spconf,amsmath,graphicx,url}
\usepackage{xcolor}

\usepackage{microtype}
\usepackage{xspace}
\newcommand{\etal}{{\it et al.}\xspace}

\definecolor{hermancolor}{HTML}{FF6600}
\definecolor{dancolor}{HTML}{9A00FF}
\definecolor{kayodecolor}{HTML}{00BFFF}

\newcommand{\config}[1]{\texttt{#1}}
\newcommand{\mylabel}[1]{{\color{gray} \footnotesize \sf #1}}

\usepackage[prependcaption,textsize=scriptsize]{todonotes}
\usepackage{combelow}

\usepackage{pifont}
\newcommand{\cmark}{\ding{51}}%
\usepackage{booktabs}
\usepackage{tabularx}
\usepackage{multirow}
\usepackage{siunitx}
\newcolumntype{C}{>{\centering\arraybackslash}X}
\newcolumntype{L}{>{\raggedright\arraybackslash}X}
\newcolumntype{R}{>{\raggedleft\arraybackslash}X}

\usepackage{tikz}
\usepackage{mathtools}
\usepackage{fontawesome}
\usetikzlibrary{backgrounds}
\usetikzlibrary{positioning}
\usetikzlibrary{fit}

\usepackage{amssymb}
\usepackage{namedtensor}
\usepackage{stmaryrd}

\ndef{\temp}{T}
\ndef{\emb}{K}
\ndef{\vocab}{W}

\newcommand{\yoruba}{Yor\`ub\'a\xspace}

\renewcommand{\paragraph}[1]{\noindent\textbf{#1}$\,$}

\usepackage{cite}  
\title{YFACC: A Yor\`ub\'a speech--image dataset for cross-lingual keyword localisation through visual grounding}

\name{
    Kayode Olaleye$^1$, Dan Onea\cb{t}\u{a}$^2$, Herman Kamper$^1$
    \thanks{%
        This work is supported in part by the National Research Foundation of South Africa (grant no. 120409)
        and by the Romanian Ministry of Education and Research (CNCS-UEFISCDI, number PN-III-P1-1.1-PD-2019-0918).%
    }
}
\address{
    $^1$Stellenbosch University, South Africa\\
    $^2$University Politehnica of Bucharest, Romania
}

\begin{document}
%
\maketitle
\begin{abstract}
Visually grounded speech (VGS) models are trained on images paired with unlabelled spoken captions. Such models could be used to build speech systems in settings where it is impossible to get labelled data, e.g.\ for documenting unwritten languages. However, most VGS studies are in English or other high-resource languages. This paper attempts to address this shortcoming. We collect and release a new single-speaker dataset of audio captions for 6k Flickr images in \yoruba---a real low-resource language spoken in Nigeria. We train an attention-based VGS model where images are automatically tagged with English visual labels and paired with \yoruba utterances. This enables cross-lingual keyword localisation: a written English query is detected and located in \yoruba speech. To quantify the effect of the smaller dataset, we compare to English systems trained on similar and more data. We hope that this new dataset will stimulate research in the use of VGS models for real low-resource languages.
\end{abstract}


\section{Introduction}
\label{sec:introduction}

\begin{figure}[!b]
    \centering
    \includegraphics[width=0.90\columnwidth]{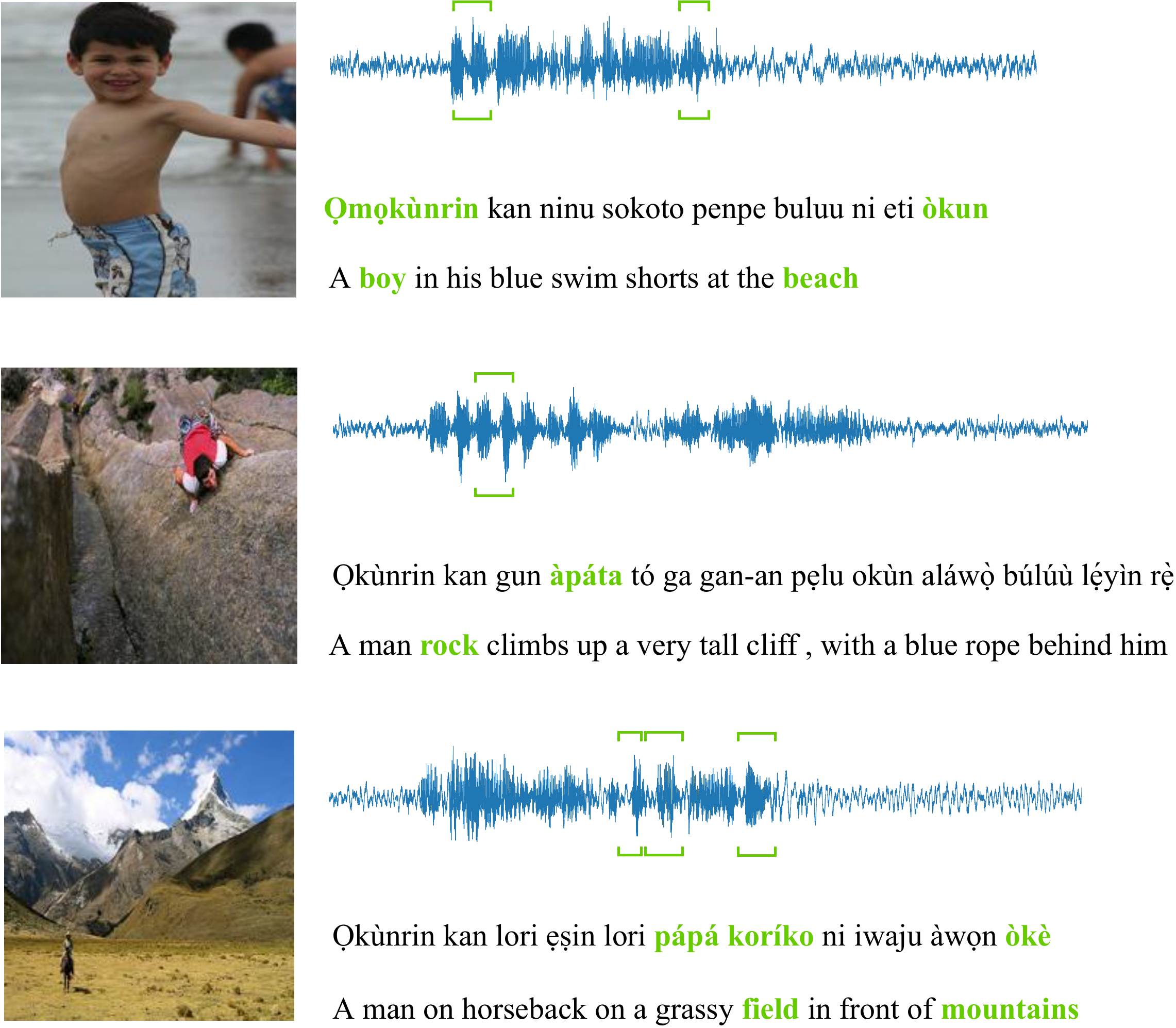}
    \caption{%
        Samples from the new 
        \yoruba Flickr Audio Caption Corpus (YFACC):
        images from the Flickr8k dataset (left) are paired with \yoruba audio captions (right), 
        which are based on the original English captions. 
        For a subset of keywords (green) we provide temporal alignments for evaluation.
    }
    \label{fig:sample_datasets}
\end{figure}

Imagine you are a linguist tasked with documenting a low-resourced language.
You collect 
	a set of recordings
from native speakers and would like to learn how 
	common words
(such as \textit{child}, \textit{water}, \textit{red}) are spoken in that new language.
To aid with this process, one would hope to use machine learning tools.
However, developing automated models requires labelled data (e.g.\ audio and transcripts), 
which are scarce or even completely unavailable (as is the case with languages that do not have a written form).
In this situation, instead, it might be reasonable to assume that we can collect images paired with their spoken description.
In this paper, we investigate 
this scenario in which we attempt to cross-lingually localise keywords in a low-resourced language (\yoruba). We assume that visual context is the only source of supervision.


Models that learn from images and their spoken captions are known as visually grounded speech (VGS) models \cite{driesen2010, synnaeve2014b, harwath2015, harwath2016,  harwath2017, harwath2018a, harwath2018b, eloff2019, harwath2019a, harwath2019b, nortje2020}.
While previous work on VGS has shown promise for performing a number of downstream tasks in low-resource conditions, very few papers validate the approach in a realistic setting---most papers use unlabelled English speech or data from another high-resource language~\cite{harwath2018b, ohishi2020} to simulate a low-resource setting.


To
address this shortcoming in VGS research,
 we introduce a new speech--image dataset in a low-resourced language, \yoruba.
The \yoruba language is one of the three official languages in Nigeria, and even though 
it is spoken natively by more than 44M speakers,
there are only a handful of high-quality datasets available in this language
\cite{vanniekerk2014a,gutkin2020,meyer2022,ogayo2022}.
Our dataset is built by augmenting the Flickr8k dataset~\cite{rashtchian2010,hodosh2015ijcai} with spoken captions: from a single \yoruba speaker, we record 6k \yoruba utterances corresponding to translations of the original English image captions.
We also provide manual alignments with start and end timestamps for 67 selected keywords in each of the 500 test utterance.
We call the resulting dataset the \yoruba Flickr Audio Captions Corpus (YFACC).
Examples of samples from YFACC are shown in Figure~\ref{fig:sample_datasets}.
A major advantage of extending Flickr8k is that it allows reusing existing annotations (English recordings \cite{harwath2015}, translations in other languages \cite{elliott2016,li2016icmr}), which could be used to construct 
richer and potentially more interesting tasks in future work.

In this paper we specifically use YFACC 
for the tasks of cross-lingual keyword detection and localisation: given an English text query, we detect whether the query occurs in \yoruba speech, and if it is detected, we localise where in the utterance the query occurs.
This is a novel scenario that resembles language documentation.
Prior work has considered cross-lingual keyword detection~\cite{kamper2018} (whether the query appears \textit{anywhere} in an utterance), but localisation was not attempted (and they considered the high-resource English--German language pair).
Other work has considered VGS-based localisation~\cite{olaleye2021,olaleye2022}, but in these monolingual studies the query keywords came from the \textit{same} language as the search utterances.
To our knowledge, we are the first to attempt cross-lingual keyword localisation; while more challenging, this task is also potentially more impactful in developing applications for linguists. 
Our approach applies the model of \cite{olaleye2022} to the cross-lingual setting:
images are automatically tagged with English visual labels, which serve as targets for an attention-based model that takes \yoruba speech as input.
This enables the \yoruba speech model to be queried with English text.

Apart from the inherent challenges in the cross-lingual retrieval task,
the English--\yoruba pair is also a challenging combination, as the two languages belong to different families (Indo-European and Nigerian-Congo) and they are associated with very different cultures. 
Through systematic experimentation, we quantify the impact of this language mismatch: we analyse
how well the selected keywords match across languages,
we compare to a monolingual English VGS model trained on similar amounts of data to YFACC,
and we transfer English speech representations to \yoruba (resulting in improved retrieval performance).

To summarise, 
our contributions include:
(i) the release of a new speech--image dataset in \yoruba,
(ii) a baseline system for cross-lingual keyword detection and localisation, and
(iii)~a multi-facet cross-lingual analysis of systems.
We hope that YFACC 
will stimulate research in the use of VGS models for solving speech tasks in real low-resource languages.
We release the data at {\small \url{https://www.kamperh.com/yfacc/}}.

\section{Related work}
\label{sec:related_work}

\begin{table*}
    \centering
    \small
    \caption{%
        A comparison of the open-source \yoruba speech corpora in terms of the available modalities 
        and quantitative characteristics. 
        The new 
        YFACC datasets provides alignments for a subset of 67 keywords.
    }
    \label{tbl:keyword_localisation_all_tasks_evaluation}
    \begin{tabular}{@{}l@{}lccccccrrrc@{}}
        \toprule
                                                  &  & \multicolumn{5}{c}{Modalities}                                         & & \multicolumn{1}{c}{Speakers}         & \multicolumn{1}{c}{Duration}        & \multicolumn{1}{c}{Utterances}       & \multicolumn{1}{c}{Sampling} \\
        \cmidrule(lr){3-7}
        Dataset                                   &  & Speech & Text (yo) & Text (en) & Images & Alignments &  & \multicolumn{1}{c}{\mylabel{number}} & \multicolumn{1}{c}{\mylabel{hours}} & \multicolumn{1}{c}{\mylabel{number}} & \multicolumn{1}{c}{\mylabel{kHz}} \\
        \midrule
        van Niekerk et al. \cite{vanniekerk2014a} &  & \cmark & \cmark    &           &        &            &  & 33                                   & 2.75                                & 4316                                 & 16 \\
        Gutkin et al. \cite{gutkin2020}           &  & \cmark & \cmark    &           &        &            &  & 36                                   & 4.00                                & 3583                                 & 48 \\
        Meyer et al. \cite{meyer2022}             &  & \cmark & \cmark    &           &        &            &  &                                      & 33.30                                & 10228                                & 48 \\
        Ogayo et al. \cite{ogayo2022}             &  & \cmark & \cmark    &           &        &            &  &                                      & 18.04                               & 10978                                & \\
        YFACC (ours)                              &  & \cmark & \cmark    & \cmark    & \cmark & partial    &  & 1                                    & 13.3                                & 6000                                 & 48 \\
        \bottomrule
    \end{tabular}
    \label{tbl:yoruba_corpus_overview}
\end{table*}

\paragraph{Visually-grounded speech models.}
There is a growing body of work considering how speech processing systems can be developed using images paired with spoken captions~\cite{synnaeve2014b, harwath2015, harwath2016, kamper2017a, harwath2018a, kamper2018, harwath2018b, harwath2019a, harwath2019b}.
One line of work trains a model to project images and speech into a joint embedding space~\cite{harwath2016,chrupala2017}.
Another line uses an external image tagger with a fixed vocabulary to obtain soft text labels, which are then used as targets for a speech network that maps speech to keyword labels~\cite{kamper2019b, pasad2019}.
Our work falls in the second category and investigates how these systems perform in a real low-resource environment.

\paragraph{Cross-lingual speech-keyword retrieval.}
Early work in cross-lingual retrieval adopted
 a direct approach of using text transcriptions for training the system instead of visual grounding.
E.g.\ \cite{sheridan1997} proposed to cascade automatic speech recognition 
with text-based cross-lingual information retrieval~\cite{oard1998}. 
However, this approach is only possible when transcribed speech is available in the target language to build a recogniser. 
Recent work has proposed models that can translate speech in one language directly to text in another~\cite{duong2016, goldwater2017, weiss2017, berard2018}, but these methods also require transcribed data: parallel speech with translated text.
\cite{kamper2018} attempted 
cross-lingual keyword spotting without the need of transcriptions by using the visual information as the link between the languages.
Our approach shares this core idea, but differs in terms of the languages used and the task tackled (as motivated in Sec.~\ref{sec:introduction}).

\paragraph{\yoruba speech corpora.}
%
%
Up until very recently, there were only two open-source \yoruba speech datasets:
the Lagos-NWU \yoruba Speech Corpus \cite{vanniekerk2014a}, consisting of approximately 2.75 hours recorded by 17 male and 16 female speakers at 16 kHz,
and the corpus of Gutkin \etal~\cite{gutkin2020}, consisting of roughly 
4 hours of 48 kHz recordings from 36 male and female volunteers. 
Concurrent work \cite{meyer2022,ogayo2022}, acknowledging the importance of creating datasets for African languages, has released more sizeable datasets geared towards building text-to-speech systems;
Table~\ref{tbl:yoruba_corpus_overview} gives their statistics.
We believe that YFACC, 
being a multimodal speech-image dataset,
will make a further useful contribution to this short list.


\section{The Yor\`ub\'a flickr audio caption corpus (YFACC)}
\label{sec:yfacc}

We introduce the new \yoruba Flickr Audio Caption Corpus (YFACC).
The YFACC dataset extends the Flickr8k image--text dataset~\cite{rashtchian2010,hodosh2015ijcai} to \yoruba with three modalities:
\yoruba translations of 6k of its captions;
corresponding spoken recordings of these translations;
temporal alignments of 67 \yoruba keywords for a subset of 500 of the captions.

\subsection{Data collection}

We 
selected a random set of 6k images from the Flickr8k dataset
and for each image we picked 
one out of the 
five associated captions.
The resulting English captions were manually translated to \yoruba by two native speakers.
Partial diacritics were added to aid the recording process.
All translations were recorded from 
a single male native speaker.
A primary motivation for collecting speech from only a single speaker is that this is representative of a real language documentation setting, where a linguist will only have access to a small number of speakers.
Recordings were made with a BOYA BY-M1DM dual omni-directional lavalier microphone using
a sampling rate of 48 kHz and two channels.

The 6k utterances were split into train (5k), development (500) and test (500) sets.
For the 500 test utterance, a native speaker produced
temporal alignments for a set of 67 \yoruba keywords corresponding to the 67 English keywords 
defined in \cite{kamper2019b}.
These keywords were selected such that they are visually groundable and provide high inter-annotator agreement,
so they are suitable to evaluate the localisation algorithm.
We prepare the alignments with \texttt{Praat}~\cite{boersma2001}. 
Alignments include the start time, end time, and duration of each of the 67 keywords.
Samples from YFACC are shown in Figure \ref{fig:sample_datasets}.

\subsection{Relation to Flickr8k and other \yoruba datasets}
\label{sec:relationship}

Our dataset is based on the Flickr8k dataset~\cite{rashtchian2010,hodosh2015ijcai},
which is a bimodal (image and text) dataset
consisting of 8k images and 40k captions (each image is annotated with a textual description by five different people).
The images (and consequently their captions) generally depict relations and actions involving people or animals.
Flickr8k was initially used for image captioning and cross-modal retrieval, but was subsequently
extended with translations \cite{elliott2016,li2016icmr}, spoken captions \cite{harwath2015}, and bounding box annotations for the visual concepts \cite{plummer2015}.
Of particular interest to us is the spoken version, Flickr Audio Caption Corpus (FACC), introduced by Harwath and Glass \cite{harwath2015}.
FACC
was obtained
using Amazon's Mechanical Turk 
to crowdsource English
audio captions for each of the text captions. 
The FACC dataset subsequently enabled visually-grounded tasks, such as image--speech alignment \cite{harwath2015}, semantic speech retrieval \cite{kamper2019b}, image-to-speech synthesis \cite{hasegawa2017} and speech-to-image generation \cite{li2020jstsp}.

Compared to previous \yoruba datasets,
ours is different in that 
it has a richer set of modalities, as shown in Table \ref{tbl:yoruba_corpus_overview}.
By extending Flickr8k we are able to reuse its existing annotations and build a multi-modal multi-lingual dataset, which, similarly to FACC, allows for a wide variety of tasks to be defined.
In this current paper we specifically consider the task of visually-grounded cross-lingual keyword localisation.
Example tasks that can be considered in the future include speech-to-speech machine translation or grounding visual concept in images from \yoruba speech.
The fact that the recordings are done by a single speaker and at high quality (48 kHz), also enables the future development of text-to-speech technology for \yoruba.


\subsection{Cross-linguistic differences}
\label{ssec:yfacc-cross-differences}


\yoruba is a very different language from English both linguistically and culturally.
In what follows we discuss some of the specific aspects
as they relate to
YFACC.

\paragraph{Tone marks.}
In \yoruba, pronunciation and writing involves using three types of tone marks \cite{wood1879}:
\textit{d\`o}, a falling tone (grave accent);
\textit{re}, a flat tone (no accent);
\textit{m\'i}, a rising tone (acute accent).
These marks are applied to the top of the seven \yoruba vowels (\textit{a, e, \d{e}, i, o, \d{o}, u}), and the four nasal vowels (\textit{\d{e}n, in, \d{o}n, un}).
The tone marks are key to understanding \yoruba, since without them many words become ambiguous.
E.g.\ 
\textit{okun} can stand for \textit{\`okun} (beach), \textit{ok\`un} (rope) or \textit{okun} (strength),
and \textit{\d{o}k\d{o}} can stand for \textit{\d{o}k\d{\`o}} (car), \textit{\d{o}k\d{\'o}} (hoe) or \textit{\d{o}k\d{o}} (husband).
    In practice including all the marks when writing is a time consuming activity.
    So for our dataset we provide diacritics
    only for the 67 keywords, 
    since only these are used for evaluation.
    However, in the future it might be possible to generate the marks for the remaining words using an automated approach~\cite{orife2020}.
    
\begin{figure}
    \centering
    \includegraphics[width=0.35\textwidth]{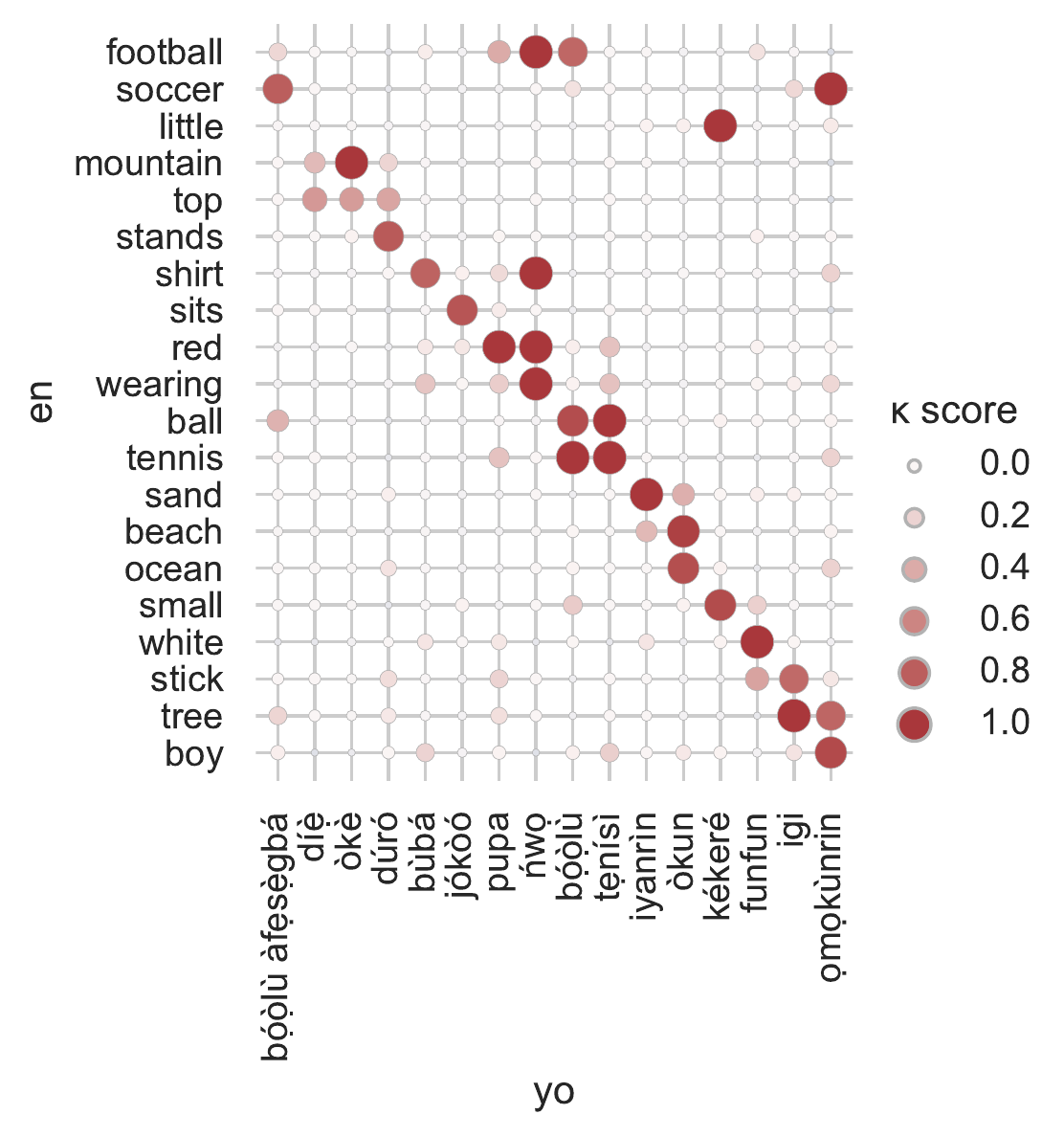}
    \caption{%
        Co-occurrences of English--\yoruba keyword pairs on the 500 test utterances in YFACC computed as normalised Cohen's $\kappa$ score.
        For readability, we pick keywords 
        from the 67-word vocabulary.
    }
    \label{fig:keyword-correlation}
\end{figure}

\paragraph{Keywords across the two languages.}
Even if YFACC is built by translating the Flickr8k
dataset,
a keyword that appears in the English caption is not guaranteed to appear in the corresponding \yoruba translation, and vice versa.
For example, 
\textit{A blonde woman appears to wait for a ride} is translated to \yoruba as \textit{Ob\`inrin bilondi nd\'ur\'o de \d{o}k\d{\`o}}
(which back-translates to \textit{A blonde woman \underline{stands} for a \underline{car}}), making the keywords \textit{stands} and \textit{car} appear in the \yoruba caption.
We quantify the common appearances of the keywords across the two languages in Figure \ref{fig:keyword-correlation}.
Intuitively, if
an English and an \yoruba word tend to co-occur (i.e., appear in the corresponding English--\yoruba translation samples, as is the case with the \textit{ride}--\textit{\d{o}k\d{\`o}} pair in the previous example),
that pair will yield a large score.
Indeed, we observe a strong correlation along the diagonal, suggesting that English keywords do correlate with their translations, but there are also some exceptions.
Some \yoruba words have multiple meanings, so they co-occur with more than a single English word,
e.g.\ \textit{\`okun}, which means both \textit{beach} and \textit{ocean}.
Conversely, some English words have multiple meanings:
\textit{football} is not just the sport, but also the \textit{ball};
\textit{little} refers not only to size, but also to age (\textit{little boy}; \textit{\d{o}m\d{o}k\`unrin k\'eker\'e}).
Finally, some correlations appear just because words co-occur together in both languages: \textit{\'nw\d{o}} (\textit{wearing}) correlates with \textit{shirt}, \textit{red}, and \textit{football}.

\paragraph{Cultural differences.}
Starting from an English dataset may induce a mismatch regarding the cultural aspects, as discussed in \cite{hershovich2022acl}.
Most of the 67 selected keywords in Flickr8k pertain to general cross-cultural concepts (such as \textit{children}, \textit{young}, \textit{swimming}),
but there are some of the English words that have no indigenous \yoruba translation.
E.g.\ words like  \textit{jacket} or \textit{tennis}
are usually translated based on their English pronunciation by adding extra marks to the English word: \textit{j\'ak\d{\`e}t\`i} and  \textit{t\d{e}n\'is\`i}. 
Perhaps the only keyword that presents little interest to the typical Nigerian is the English word \textit{skateboard}.

\section{Cross-lingual keyword localisation with a visually-grounded speech model}
\label{sec:cross_lingual_tasks}


\begin{figure}[t]
    \centering
    \includegraphics[width=0.99\columnwidth]{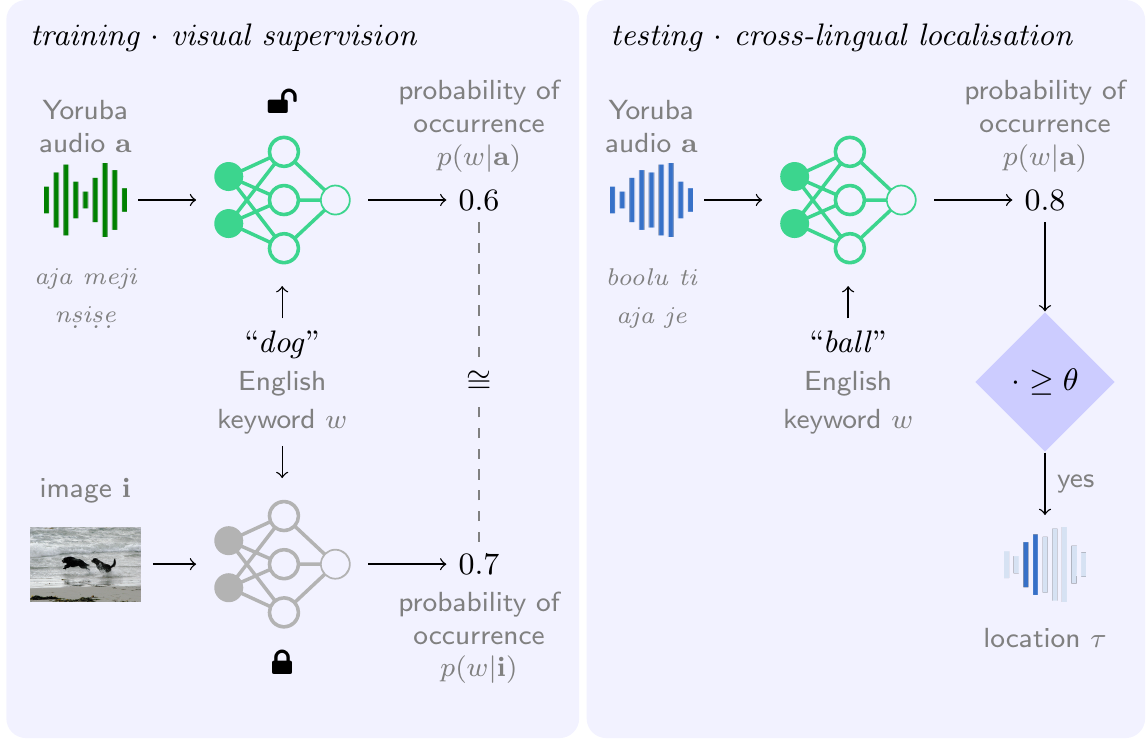}
    \caption{%
        Cross-lingual keyword localisation with a visually-grounded speech model.
        \textit{Left:} At train time, the speech model is trained to predict whether an English keyword occurs in a \yoruba audio utterance based on the visual supervision of a pretrained and frozen (English) image model.
        \textit{Right:} At test time, given an audio utterance in \yoruba, the model outputs the probability of an English keyword occurring in the utterance.
        If this probability is over a predefined threshold $\theta$, then the model additionally locates the keyword in the utterance based on the attention weights in the network.
    }
    \label{fig:overview}
\end{figure}

We specifically consider
the task of cross-lingual keyword localisation:
given an English keyword and a spoken utterance in \yoruba, 
we want to automatically find the location in the audio of the \yoruba word that corresponds to the given English keyword.
If the keyword does not occur, the model should indicate that the keyword was not detected.
Crucially, we assume that the only source of supervision comes from a visual channel.
Concretely
in our case we assume we have access to a set of images associated with corresponding audio captions (YFACC).
In order to be able use this weak source of supervision, we follow 
an almost identical approach to the monolingual VGS-based localisation method of
\cite{olaleye2022}, which we adapt to the cross-lingual setting.

\subsection{Training: Visual supervision}

Our model consists of an audio network that processes \yoruba 
speech and outputs the probability that a given English keyword occurs (anywhere) in the utterance.
This audio model is trained by cross-modal supervision: 
it learns to reproduce the predictions of a pretrained image network on the associated image for a predefined set of visual categories.
The weights of the audio network are optimised through the binary cross-entropy loss between the image and audio predictions;
the loss is averaged over the keywords in the vocabulary.
The categories are defined on English keywords (such as \textit{dog}, \textit{child}, \textit{ball}), but they also correspond to \yoruba words which occur in the audio,
with the visual channel linking the two languages.
See Figure~\ref{fig:overview} (left) for an 
illustration of the training process.

\subsection{Testing: Cross-lingual localisation}

At test time, given a \yoruba utterance---no images are needed for inference---our model outputs the probability of occurrence of a given English keyword from the visual vocabulary.
When this probability is over a certain threshold $\theta$, we assume that the keyword occurs in the utterance.
Once a keyword is detected, we also want to
locate the corresponding \yoruba word.
To obtain the time location of the Yoruba word, we leverage the extra structure of our model, namely its attention weights over the input utterance.
We pick as the most probable location of the input keyword the time stamp with the highest 
weight.
See Figure~\ref{fig:overview} (right) for an illustration of the testing phase.

\subsection{Architecture: Attention network}

The audio network consists of four subcomponents: an audio encoder, a keyword encoder, an attention layer and a classification network.
The audio encoder---a convolutional network in our case---embeds the \yoruba audio to a sequence of feature vectors,
while the keyword encoder assigns an embedding vector to the English keyword using a learnable lookup table.
The audio features are pooled temporally into a single vector based on the keyword embedding using the attention layer.
Finally, the aggregated feature vector is then projected to a classification score using a small multi-layer perceptron network.
As mentioned in the previous subsection, we use the location of the maximum attention scores as the location prediction of the query keyword.
This architecture is based on the work of \cite{olaleye2022}, which in turn draws inspiration from~\cite{tamer2020},
an attention-based graph-convolutional network for visual keyword spotting in sign language videos.

\section{Experimental setup}
\label{sec:exp_setup}

\subsection{Datasets}

In our experiments, apart
from the new 
YFACC dataset (Sec.~\ref{sec:yfacc}) we also use the previous English FACC dataset (Sec.~\ref{sec:relationship})
\cite{harwath2015} for the monolingual English VGS systems which we compare to. 
We experiment with two variants of FACC that differ in the amount of training data:
\config{en-30k}, which contains all the original 30k training samples, and
\config{en-5k}, which contains a subset of 5k training samples to match the size of
YFACC.
The test data used for models trained on FACC are the 500 English utterances corresponding to the YFACC test set.


\subsection{Model architecture and implementation details}

We use the same architecture as in~\cite{olaleye2022}, but carry out a hyperparameter search over the learning rate, activations functions and size of last layer.
This search is done based on
F1 keyword detection score on the development set of FACC.
Compared to the original architecture in~\cite{olaleye2022}, the only change is increasing the last layer's size from 4096 to 8192; this improved development-set F1 
from 14.4\% to 19.4\%.

We downsample YFACC to 16 kHz to match FACC. All utterances are
encoded as mel-frequency cepstral coefficients, 
which are further augmented at train time using SpecAugment~\cite{park2019}.
The model's visual targets are computed with a VGG-16 network~\cite{simonyan2014} trained on MSCOCO~\cite{lin2014} for 67 keywords as described in \cite{kamper2019b}.
Our VGS models
are optimised with Adam~\cite{kingma2015} using a learning rate of ${10}^{-4}$. 
The PyTorch implementation is publicly available.%
\footnote{\url{https://github.com/kayodeolaleye/keyword_localisation_speech}}

\subsection{Evaluation of cross-lingual keyword localisation}
\label{sec:eval}

Our main metric is precision, which is computed as the ratio of true positive samples to retrieved samples.
A sample is \textit{retrieved} if the detection score for the given English keyword is greater than a threshold $\theta$.
A sample is a \textit{true positive} if
(1)~it is retrieved and 
(2)~the given English keyword is correctly localised 
in the \yoruba utterance,
i.e., the predicted location falls within the interval of the corresponding spoken \yoruba keyword.
We refer to this metric as \textit{actual keyword localisation}~\cite{olaleye2022}.
For all experiments, we set $\theta$ to $0.5$, which gave best performance on the FACC development set. 

\section{Experimental results and analysis}
\label{sec:exp_results}

\begin{table*}
    \centering
    \caption{
        Results of visually-grounded speech models
        on cross-lingual tasks (querying English in \yoruba audio) on the proposed YFACC dataset (first three rows) and
        on monolingual tasks (querying English in English audio) on the FACC dataset (last two rows).
        The models are evaluated for the main task of keyword localisation (actual localisation) and two other easier tasks (oracle localisation and keyword detection), which provide upper bounds on the actual localisation performance.
        The experiments are repeated three times by changing the random seed and we report the mean performance and its standard deviation.
    }
    \begin{tabular}{@{}lcc@{\hspace{0.75\tabcolsep}}c@{\hspace{0.75\tabcolsep}}ccrccrrr}
        \toprule
                            &  &                    &  &                    &  &                                          &       &  & \multicolumn{1}{c}{Actual}              & \multicolumn{1}{c}{Oracle}             & \multicolumn{1}{c}{Keyword}   \\
                            &  & Keyword            &  & Audio              &  & \multicolumn{1}{c}{Training}             & Initialisation &  & \multicolumn{1}{c}{localisation}        & \multicolumn{1}{c}{localisation}       & \multicolumn{1}{c}{detection} \\
                            &  & \mylabel{language} &  & \mylabel{language} &  & \multicolumn{1}{c}{\mylabel{num. utts.}} &       &  & \multicolumn{1}{c}{\mylabel{precision}} & \multicolumn{1}{c}{\mylabel{accuracy}} & \multicolumn{1}{c}{\mylabel{precision}} \\
        \midrule
        \config{random}     &  & \multicolumn{6}{l}{Random baseline}                                              &  &           $0.1 \pm 0.0$ &           $5.3 \pm 0.9$ &           $2.4 \pm 0.2$ \\
        \config{yo-5k}      &  & \multicolumn{1}{l}{English} & $\rightarrow$ & \yoruba &  & 5k  & random          &  &          $16.0 \pm 1.6$ &          $23.1 \pm 3.0$ &          $26.7 \pm 6.4$ \\
        \config{yo-5k-init} &  & \multicolumn{1}{l}{English} & $\rightarrow$ & \yoruba &  & 5k  & \config{en-30k} &  & $\mathbf{22.8} \pm 4.3$ & $\mathbf{35.0} \pm 2.5$ & $\mathbf{33.3} \pm 7.1$ \\
        \midrule
        \config{en-5k}      &  & \multicolumn{1}{l}{English} & $\rightarrow$ & English &  & 5k  & random          &  &          $21.3 \pm 4.4$ &          $27.3 \pm 2.8$ &          $29.5 \pm 5.0$ \\
        \config{en-30k}     &  & \multicolumn{1}{l}{English} & $\rightarrow$ & English &  & 30k & random          &  & $\mathbf{26.9} \pm 1.8$ & $\mathbf{44.0} \pm 0.9$ & $\mathbf{39.2} \pm 1.7$ \\
        \bottomrule
    \end{tabular}
    \label{tbl:main-results}
\end{table*}




Our goal is to
investigate to what extent cross-lingual keyword localisation is possible with a VGS 
model trained in an extreme low-resource setting.
We specifically consider the case where we have 
only 13.3 hours of untranscribed \yoruba speech with corresponding images.


English--\yoruba cross-lingual results are given at the top of Table~\ref{tbl:main-results}.
The 
cross-lingual model \config{yo-5k} obtains a performance of about $16.0$\% for our main task, actual keyword localisation (Sec.~\ref{sec:eval}).
Although this 
appears modest when viewed in isolation, it is much better than the performance of a random model ($0.1$\%).
More
importantly, the result 
falls within the range of results ($21.3\pm4.4$\%)
that are obtained in a monolingual setting (querying English keywords in English speech) with a model trained on equal number of English utterances (\config{en-5k}).
It can be argued that the cross-lingual setting is more challenging because the visual keywords are more English-centric and do not account for the cultural and linguistic differences to \yoruba (Sec.~\ref{ssec:yfacc-cross-differences}).
On the other hand, YFACC is easier in the sense its a single-speaker dataset, whereas the FACC dataset involves multiple speakers.
Next we consider aspects that can help improve performance.

\paragraph{Amount of training data.}
While for the cross-lingual setting we are limited to the 5k training utterances available in 
YFACC, 
we can extrapolate the performance on more data based on corresponding monolingual setting using 
FACC.
In Table~\ref{tbl:main-results} we see that training with six times more data, 
unsurprisingly, yields considerable improvements, from $21.3$\% (\config{en-5k}) to $26.9$\% (\config{en-30k}).
However, while increasing the training size is the most straightforward way to improve performance, it might also be the most challenging in practice.

\paragraph{Speech representations.}
Training a model from scratch in low-resource conditions is challenging.
A common practice is to start from a pretrained model and to then fine-tune it 
on the downstream task.
Here we do this by initialising
the training on \yoruba from a VGS model trained on all the image--English audio pairs in FACC.
Such English pairs
are arguably more readily available than 
\yoruba utterances. 
We observe a strong boost in performance---from $16.0$\% (\config{yo-5k}) to $22.8$\% (\config{yo-5k-init}). 

\paragraph{Decoupling the tasks.}
Keyword localisation involves two steps:
(1) detecting whether the query keyword appears (somewhere) in the utterance, and, if so,
(2) localising the keyword in the given utterance.
In order to understand which component of 
our approach
requires 
improvements, it is instructive to independently evaluate each of these steps.
To do that, we assume that one of the steps is perfect, resulting in 
two new tasks:
oracle localisation (assumes that detection is perfect) and keyword detection (assumes that localisation is perfect).
Whilst not directly comparable, the results for the two tasks (last two columns in Table~\ref{tbl:main-results}) indicate that both components can be further improved.
Also note 
that the previous conclusions in terms of relative performance between the models still hold 
on these 
new tasks. 

\paragraph{Individual keyword performance.}
The performance varies widely across the 67 keywords in the vocabulary.
We observe that there are keywords with good performance (better than the average), such as 
\textit{brown} (\textit{b\'ur\'a\`un}; $100$\%),
\textit{bike} (\textit{k\d{è}k\'e}; $94.1$\%) or
\textit{grass} (\textit{kor\'iko} $90.9$\%).
But there are many others on which the model struggles.
The reason for the latter include poor visual grounding (e.g., \textit{camera}, \textit{wearing}, \textit{large}) and confusion between semantically related concepts (\textit{riding} often retrieves \textit{k\d{è}k\d{é}}, i.e. \textit{bicycle}; \textit{swimming} retrieves \textit{odò adágún}, i.e. \textit{pool}). 
Paying close attention to the particularities of the individual keywords can serve as inspiration for
future improvements. 

\begin{figure}[!t]
    \centering
    \includegraphics[width=0.99\linewidth]{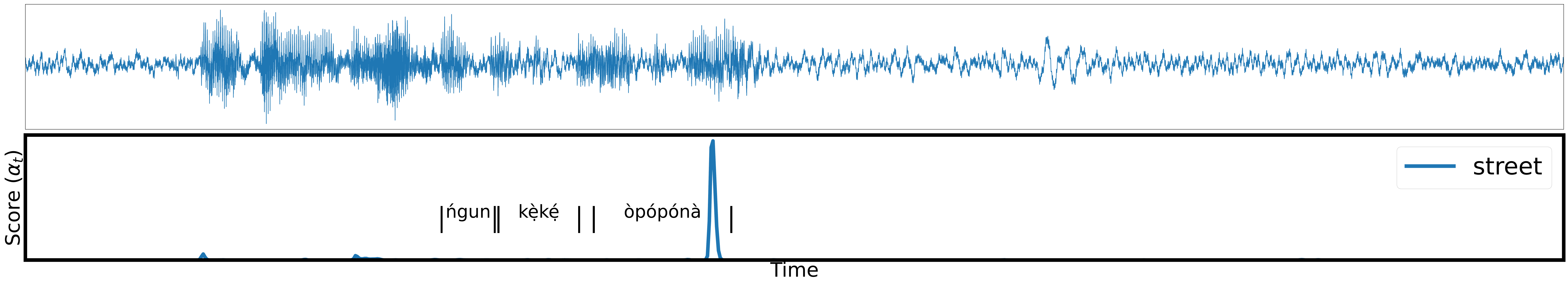} \\[-3pt]
    {\footnotesize (a)} \\[5pt]
    \includegraphics[width=0.99\linewidth]{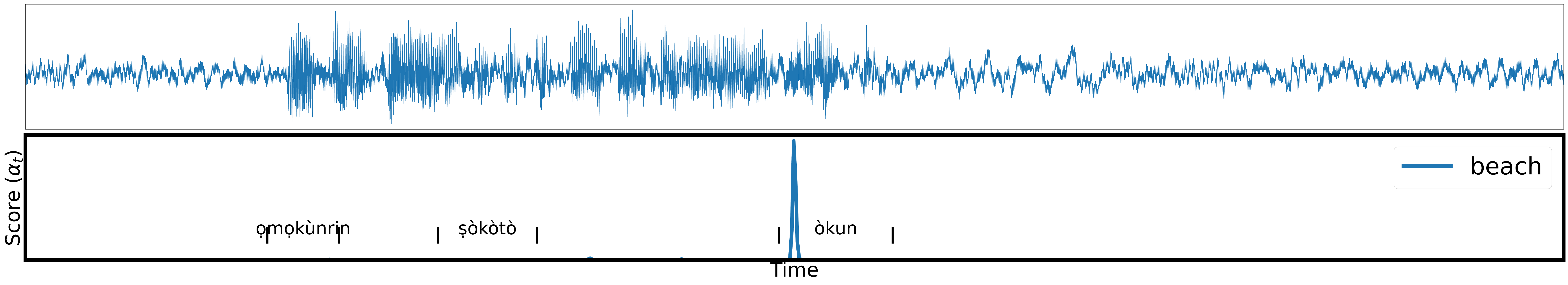} \\[-3pt]
    {\footnotesize (b)}
    \includegraphics[width=0.99\linewidth]{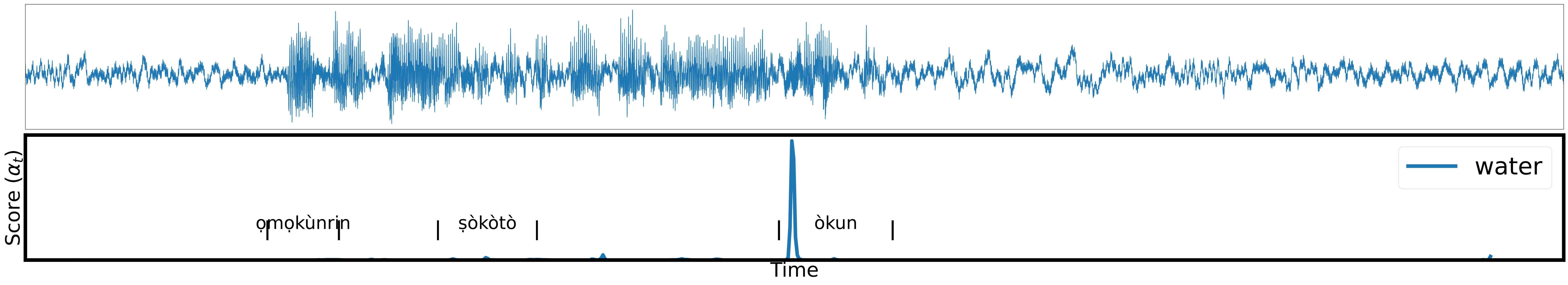} \\[-3pt]
    {\footnotesize (c)}
    \includegraphics[width=0.99\linewidth]{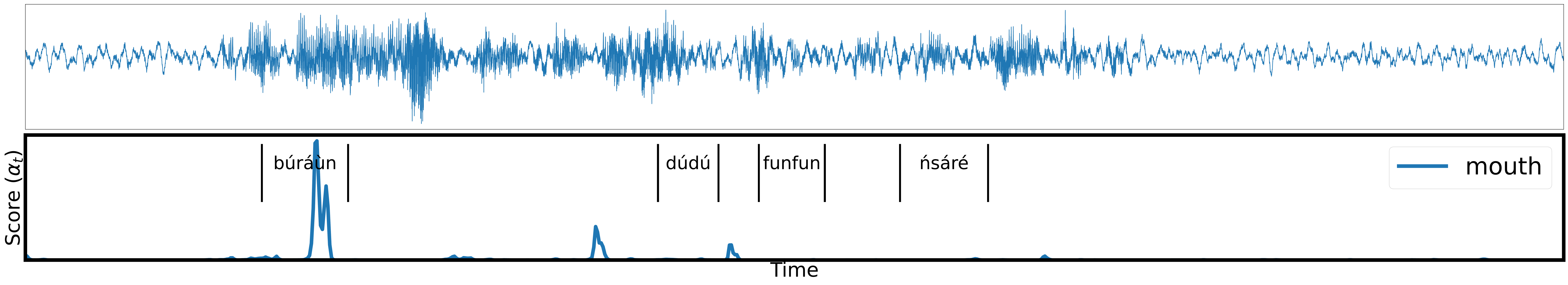} \\[-3pt]
    {\footnotesize (d)} 
   	\vspace*{-5pt}
    \caption{Examples of cross-lingual localisation with the \config{yo-5k} VGS model. The English query keyword is shown on the right.}
    \label{fig:qualitative_eval}
\end{figure}

\paragraph{Qualitative results.}
To better understand the failure modes of the cross-lingual model, we 
qualitatively assess the results.
Figure~\ref{fig:qualitative_eval}(a) and (b) show correct localisations of English keywords in \yoruba utterances:
the model fires within the frames containing the spoken word \textit{\`op\'op\'on\`a} when prompted with the English keyword \textit{street} (\ref{fig:qualitative_eval}a) and
fires within the frames of \textit{\`okun} when prompted with the keyword \textit{ocean}.
Figures~\ref{fig:qualitative_eval}(c) and (d) show 
failure cases. 
Figure~\ref{fig:qualitative_eval}(c) shows that the model also sometimes localises a semantically related word in the utterance: the spoken \yoruba word \textit{\`okun} (\textit{ocean}) is assigned the highest localisation score when prompted with the English keyword \textit{water}.
Figure~\ref{fig:qualitative_eval}(d) is 
an outright failure.

\section{Conclusion}
\label{sec:conclusions}

We introduced a new single-speaker dataset of \yoruba spoken captions for Flickr images. 
To showcase what is possible with such a small dataset (13.3 hours, 6k speech--image pairs) in a real low-resource setting, we used the data for cross-lingual keyword localisation using a visually grounded speech model, trained without any transcribed audio.
By initialising the model with an English model trained on more data, we were also able to obtain some improvements in keyword localisation and detection performance.
This dataset with the accompanying baselines will allow researchers to evaluate their visually grounded speech models in a real low-resource~setting.



\bibliographystyle{IEEEbib}
\bibliography{strings,refs}

\end{document}